\documentclass{article}

\usepackage[colorlinks=true]{hyperref}  
\usepackage{authblk}
\usepackage{newtxtext}
\usepackage{geometry}
\usepackage{mathptmx}
\usepackage{soul}\setuldepth{article}
\usepackage[table, dvipsnames]{xcolor}
\usepackage[skip=5pt plus1pt, indent=20pt]{parskip}
\usepackage[url=false,doi=false, isbn=false]{biblatex} 
\usepackage{tabularx}
\usepackage{adjustbox}
\usepackage{float}
\usepackage{lscape} 
\usepackage{array}
\usepackage{graphicx}
\usepackage{booktabs} 
\usepackage{changepage}
\usepackage{arydshln}
\usepackage{color, colortbl}
\usepackage{pdflscape}
\usepackage{afterpage}
\usepackage{array}
\usepackage[bottom]{footmisc}

\newcommand{\wrt}{{\it w.r.t. }}    
\newcommand{\eg}{\emph{e.g.}, }     
\newcommand{\ie}{\emph{i.e.}, }     
\newcommand\etc{\emph{etc.}}

\def\hb{\hbox to 11.5 cm{}}

\definecolor{lavendergray}{rgb}{0.77, 0.76, 0.82}
\colorlet{lightergray}{lightgray!40}
\newcolumntype{P}[1]{>{\raggedright\arraybackslash}p{#1}}

\author[a]{Lucia Urcelay $^*$}
\author[a]{Daniel Hinjos $^*$}
\author[a]{Pablo A. Martin-Torres $^*$}
\author[a]{Marta Gonzalez $^*$}
\author[c]{Marta Mendez}
\author[c]{Salva Cívico}
\author[a,b]{Sergio Álvarez-Napagao}
\author[a,b]{Dario Garcia-Gasulla}
\date{}

\affil[a]{Barcelona Supercomputing Center}
\affil[b]{Universitat Politécnica de Catalunya - Barcelona TECH}
\affil[c]{Hospital Clínic de Barcelona}

\title{Exploring the Role of Explainability in AI-Assisted Embryo Selection}

\begin{document}
\def\thefootnote{*}\footnotetext{Equal contributors.}

\maketitle
\begin{abstract}
In Vitro Fertilization is among the most widespread treatments for infertility. One of its main challenges is the evaluation and selection of embryo for implantation, a process with large inter- and intra-clinician variability. Deep learning based methods are gaining attention, but their opaque nature compromises their acceptance in the clinical context, where transparency in the decision making is key. In this paper we analyze the current work in the explainability of AI-assisted embryo analysis models, identifying the limitations. We also discuss how these models could be integrated in the clinical context as decision support systems, considering the needs of clinicians and patients. Finally, we propose guidelines for the sake of increasing interpretability and trustworthiness, pushing this technology forward towards established clinical practice.
\end{abstract}

\section{Introduction}\label{sec:intro}

Infertility is a common reproductive health problem that affects millions of people worldwide, causing social, psychological, physical and economic distress to the ones seeking to conceive \cite{boivin2007international}. In the coming years infertility rates are projected to grow due to environmental and lifestyle factors \cite{gore_edc-2_2015,segal_before_2019}. 
In vitro fertilization (IVF) technology is used to overcome infertility, it involves the fertilization of an egg with sperm in the laboratory, followed by the transfer of the resulting embryos into the patient's uterus. The main challenge of IVF is the selection of the embryos that will be either selected for implantation, frozen (for later implantation) or discarded (if they exhibit undesirable features). This selection is to be performed during the early hours after embryo insemination, typically between three and five days after. During this time, embryos are monitored in time-lapse imaging incubators (TLI), facilitating uninterrupted embryo growth within stable culture conditions. This technology offers a dynamic perspective on in vitro embryonic development, augmenting the clinical effectiveness of IVF \cite{mio2008time}. 

To assess quality, embryologists evaluate different morphological characteristics depending on the embryo development phase. Early development (days one to three) focus on cell number, symmetry and fragmentation rate, while embryos reaching bastocyst stage (day five) are further assessed by their expansion grade and the appearance of the inner cell mass (ICM) and the Zona Pellucida (ZP), as well as to the trophectoderm cells (TE) \cite{nasiri2015overview}.
These morphological features represent the foundation of current development assessment guidelines such as Gardner's \cite{gardner_assessment_2003} or ASEBIR in Spain. 
These approaches are limited by the subjective assessment of embryologists, which causes inter and intra-observer variability. The success rates of IVF are restricted due to these limitations. According to \cite{adamson_international_2018}, the global life birth rate via IVF in 2015 stood at 19.2\% per oocyte retrieval with fresh embryo transfer and 24.8\% with frozen embryo transfer.

Artificial Intelligence (AI), and specially Deep Learning (DL), due to its capacity for dealing with images, have recently been considered to assist in the embryo assessment and selection process. AI has the potential to facilitate and improve the process of embryo selection, increasing the implantation success rates, and reducing the chances of multiple pregnancies. AI can also mitigate inter and intra-observer variability, making results more reproducible and comparable \cite{rosenwaks2020artificial}.
Finally, AI can help reduce the financial, physical and emotional burden on patients, by optimising the treatment plan and minimising the need for repeated cycles of IVF.

Whenever AI is considered for an application which directly affects human life, trustworthiness must be assured. In this paper we review the current work on AI-assisted embryo selection with special focus on Explainable Artificial Intelligence (XAI), and how this is deployed. We analyze the limitations of current work and propose approaches that would yield AI systems that can provide better explanations for their recommendations. Moreover, we discuss key aspects in the integration of these systems in a clinical context where both clinicians and patients are involved in the decision-making process.

\section{Related Work}

This work discusses the integration of explainability components into AI solutions for embryo analysis in IVF, as well as the particularities of their integration into a human-centric decision making process. Previous solutions proposed in the bibliography are reviewed within the main discussion of this work (see \S\ref{subsec:xai}), as it contextualizes our contribution. In contrast, the related work presented in this section comprehends previous analysis and surveys conducted in the field, which have motivated this paper.

Among recent surveys on AI-assisted embryo selection in IVF, most restrict themselves to the technicalities of the machine learning problem itself. That is, what is the problem to solve, which data is used and how, and which are the top-of-the-line performance metrics achieved by the models built. A few discuss the biological aspects of the problem with particular depth \cite{kim_non-invasive_2022}, while others focus on the variety of tasks one can tackle (\eg embryo segmentation, quality grading, etc.) \cite{isa_image_2023}. The most ambitious tasks, those of predicting clinical pregnancy or fetal heartbeat from five days old fertilized embryo images, are reviewed in \cite{konstantinos_sfakianoudis_reporting_nodate}, together with the ploidy status (number of chromosomes) prediction task. \cite{louis_review_2021} includes the annotation of data (\eg development phase, cell counting) in its review, while the particularities of the time-lapse imaging are the focus of \cite{lundin_time-lapse_nodate}. On top of embryos, \cite{dimitriadis_artificial_2022} also reviews AI applications for spermatozoa and oocyte analysis. \cite{fernandez_artificial_2020} goes beyond DL methods, and also consider Bayesian Networks and SVMs in their study.
Remarkably, very few works  \cite{afnan_ethical_2021, tamir_artificial_2022, rolfes_artificial_2023} study the ethical, socio-economic, legal and cultural (ELSEC) aspects in AI-assisted embryo selection. 

None of the previously discussed works includes a review on the XAI methods used for embryo analysis, or discussion on practical and responsible deployment. This work aims at filling that gap, considering that both these issues need to be properly assessed before clinicians can integrate AI in their day-to-day practice. 

\section{The Role of Explainable Artificial Intelligence in Embryo Analysis}

Embryo analysis post-fertilization lasts for a maximum of 7 days. During that time, clinicians assess the status and evolution of embryos, as these evolve under controlled conditions. Through continuous visual monitoring and tracking of their morphological features, experts eventually discard, freeze or select embryos for transfer. Current embryo incubators include periodic microscopy, with images being recorded at frequency in the order of minutes (\eg ten) at several focal planes.

\begin{figure}[H]
  \centering
  \includegraphics[width=0.52\textwidth]{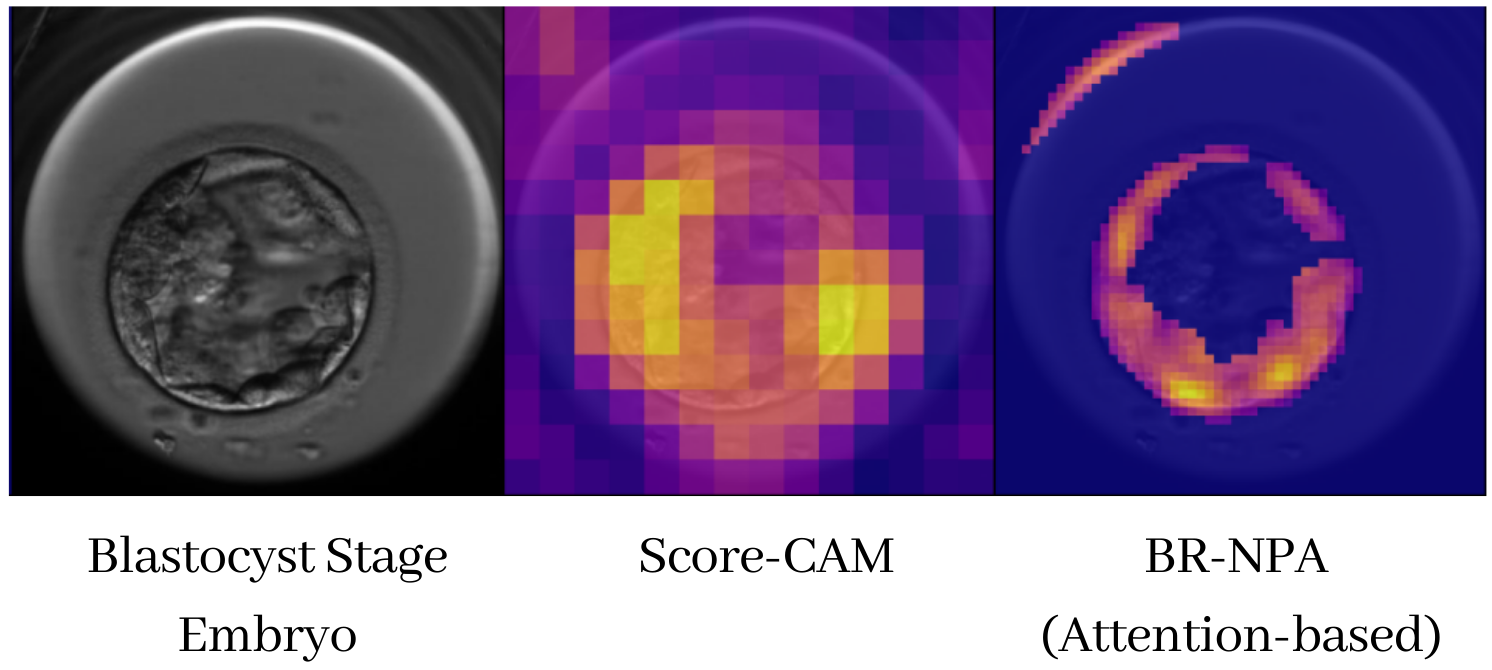}
  \caption{Saliency maps for Score-CAM and BR-NPA XAI methods from \cite{gomez_comparison_2022}.}
  \label{fig:xai_methods}
\end{figure}

AI models trained on time-lapse embryo images have shown promise in embryo analysis. Within the domain of images, explainable AI methods are often feature scorers, representing their output as saliency maps on the input image (see Figure \ref{fig:xai_methods}). By nature, XAI methods for DL are approximations to the model's real behavior. It is worth noting that XAI is critical to ensure that the decision-making process of an algorithm is as transparent and understandable to clinicians and patients as possible. The lack of explainability can lead to a lack of trust in the technology, which ultimately hinders its adoption.

In this section we review the use of XAI on embryo images. Among the methods found in the literature, we can differentiate between those applied after the model is trained (post-hoc), and those where the explainability is part of the model architecture and design (intrinsic). Post-hoc methods include model-agnostic ones such as LIME \cite{lime} and KernelSHAP \cite{shap}, that obtain the explanations by perturbing the inputs and observing the changes caused on the outputs; also, model-specific methods such as Grad-CAM \cite{gradcam} and Deep SHAP \cite{shap}, that use the parameters of neural network models to obtain the saliency maps.  Among intrinsic methods, we can find attention-based CNNs, where the explainability is obtained from attention layers. A different approach is the use of Speeded Up Robust Features (SURF) \cite{SURF} to extract local visual features, in conjunction with Gaussian Mixture Models (GMM) to obtain a Fisher vector per image \cite{kallipolitis_explainable_2022}.

\vspace{0.25cm}

\begin{table}[!]
    \centering
    \begin{tabular}{c|P{3cm}P{2.5cm}P{2.5cm}P{1.5cm}}
        \toprule
        \textbf{Ref.} & \textbf{Task} & \textbf{Architecture} & \textbf{XAI Method(s)} & \textbf{ClinicalEval} \\
        \midrule
        \cite{diakiw_artificial_2022} & Embryo Quality Grading & DenseNet, ResNet & Grad-CAM++ & No\\
        \rowcolor{lightergray}
        \cite{wu_lwma-net_2022} & Embryo Quality Grading & LWMA-Net & Grad-CAM & No\\
        \cite{kallipolitis_explainable_2022} & Embryo Quality Grading & SURF and GMM, ResNet-101, EfficientNet-B1 & SURF, Grad-CAM & No\\
        \rowcolor{lightergray}
        \cite{arsalan_human_2022} & Embryo Segmentation & MASS-Net & Grad-CAM & No \\
        \cite{sharma_explainable_2022} & Development Stage Identification & ResNet34, VGG16 & Grad-CAM, SHAP, LIME & Yes \\
        \rowcolor{lightergray}
        \cite{paya_automatic_2022} & Embryo Quality Grading & ResNet50, VGG-16 & Grad-CAM & No\\
        
        \cite{gomez_comparison_2022} & Development Stage Identification & ResNet50, ABN & B-CNN, ABN, InterByParts, Grad-CAM++, RISE, Score-CAM, Ablation-CAM, AM  & No\\
        \rowcolor{lightergray}
        \cite{enatsu_novel_2022} & Fetal Heart Pregnancy, Live Birth & ResNet18, RandomForest & Grad-CAM, SHAP  & No \\
        \cite{sawada_evaluation_2021} & Live Birth & ABN & ABN  &  Yes \\
        \rowcolor{lightergray}
        \cite{wang_deep_2021} & Embryo Quality Grading &  VGG-16  & Grad-CAM & No\\
        \cite{thirumalaraju_evaluation_2020} & Embryo Quality Grading & Xception, Inception-ResNET-v2 & CAM & No\\
        \bottomrule

    \end{tabular}
        \vspace{0.8em}
    \caption{Summary table of studies that use XAI on embryo selection. The Clinical Evaluation column indicates whether the resulting saliency maps have been evaluated by experts or not.}
    \label{table:xai_papers_mini}
\end{table}

\subsection{Review on Current Work}\label{subsec:xai}

Depending on the specific task, XAI is used and interpreted differently:

\emph{Embryo Quality Grading:} the classification of embryos (\eg day 5 blastocysts), into classes for selecting the best for implantation. This challenge is the most addressed regarding XAI. Some studies apply Grad-CAM for visualization \cite{paya_automatic_2022,wang_deep_2021} and observe that key regions that the model relies on seem to be consistent with clinical interpretation. \cite{diakiw_artificial_2022} uses Grad-CAM++ for visualization, while statistically correlating their model outputs with the Gardner score. \cite{thirumalaraju_evaluation_2020} directly use CAM saliency maps, and observe their model focus on known features like cellular fragmentation, blastomeres or vacuoles. A different approach compares Grad-CAM and SURF (computing a single Fisher Vector per image), finding the former more prone to erroneously focus on irrelevant areas \cite{kallipolitis_explainable_2022}.

\emph{Embryo Development Stage Identification:} consists on the automated labelling of the embryo stage. Challenges include deformed cell shapes, poor visual features or similarities between embryos at different stages. \cite{sharma_explainable_2022} compares Grad-CAM and LIME classification activations, concluding that LIME explanations seem less consistent with biologically important regions. Authors also suggest SHAP could be useful to identify reasons for misclassification between adjacent cleavage stages.

\emph{Blastocyst Segmentation:} four crucial morphological regions to segment: TE, ZP, ICM and BL (see \S\ref{sec:intro}). A multiscaling architecture that outputs segmentation masks for the four regions is introduced in \cite{arsalan_human_2022}, where authors apply Grad-CAM to different layers to show the evolution of activations. A U-Net has also been used to separate the background from the blastocyst \cite{kallipolitis_explainable_2022}, using an ellipse on top of the segmentation
mask to separate inner cell mass from trophectoderm.

\emph{Fetal Heart Pregnancy:}
is defined as the presence of fetal heartbeats in the uterus. Of course, the embryo implantation process does not guarantee a successful pregnancy, as it depends on many other factors (\eg age, progesterone levels, \etc). \cite{enatsu_novel_2022} uses Grad-CAM to detect morphological indicators, and SHAP to account for relevant metadata that influences the model's decision, concluding that embryo images are the best predictor for fetal heart beat, followed by age and pregnancy history.

\emph{Live Birth:} prediction from time-lapse imaging, \cite{sawada_evaluation_2021} uses an Attention Branch Network (ABN) \cite{fukui_attention_2019}. Authors visualyze embryo features of relevance for \textit{Live Birth} by exploring the attention mechanism. Results indicated that no common visual features could be associated to the predicted outcome of live or non-live birth.

To the best of our knowledge, other very important challenges in this field, such as \textit{Blastocyst Prediction} or \textit{Ploidy Detection}, are not addressed in the context of explainability. This highlights the need for further research in these tasks.

\subsection{Limitations and Suggestions}

A limited amount of research comprehends the use of AI-assisted IVF in combination with XAI methods. A representative list can be found in Table \ref{table:xai_papers_mini}. In many cases XAI is only considered through the illustration and minor discussion of a few saliency maps \cite{arsalan_human_2022, kallipolitis_explainable_2022, thirumalaraju_evaluation_2020, wu_lwma-net_2022}. Others take one more step and analyze the map activations in order to correlate them to morphological features \cite{sharma_explainable_2022, wang_deep_2021} or to the objective \cite{enatsu_novel_2022}. Few 
share the saliency maps with expert embryologists for evaluation \cite{sharma_explainable_2022, sawada_evaluation_2021}. Meanwhile, several studies conclude that saliency maps should not be used as the sole source of explainability in high risk medical domains \cite{arun_assessing_2021, ghassemi_false_2021}.

Considering the features that clinicians base their decisions on, and the difficulty of interpreting saliency maps, we suggest the use of expert models. Different models targeting different morphokinetical features, such as fragmentation score, ICM and TE grades, ZP and Pronucleus characteristics, and phase transition times. The ensemble of all these experts should be a white box model (\eg GLM, Decision Tree) to gain an intermediate level of explainability on how each feature contributes to the objective. 

In the papers reviewed, the XAI method used for interpretability is rarely baselined or justified.
Two studies compare two explainability methods, \cite{sharma_explainable_2022} (LIME vs Grad-CAM) and \cite{kallipolitis_explainable_2022} (Bag-of-visual-words-based approach vs Grad-CAM). Only one thorough comparative study was found \cite{gomez_comparison_2022}, in which nine different XAI methods, including five post-hoc ones, were evaluated using seven faithfulness metrics. This is, as well, the only study in which explainability methods are assessed through quality metrics such as the Increase in Confidence or the Average Drop \cite{chattopadhay2018grad}. 

Another limitation in the field is privacy. None of the previous works makes the code (except for \cite{arsalan_human_2022}) or the data (except for \cite{gomez_comparison_2022}) publicly available, making reproducibility impossible and replicability impractical. These flaws, in a task with a high degree of subjectivity (inter and intra-observer variability) induce a high risk of confirmation bias. This issue could be mitigated through a quantitative benchmark measuring the model's attention to certain embryo regions of interest across a relatively big amount of samples. That would require an additional segmentation model such as \cite{arsalan_human_2022}, which isolates the TE, ZP, ICM, and BL regions. A good explanation from a good model should overlap with the key segmented areas, and could be assessed with metrics like IoU. This approach is limited to blastocyst images, and subject to the accuracy of the segmentation model.

\section{Integrating Explainable Artificial Intelligence in Decision Support Systems}

The performance of AI methods on top-of-the-line metrics (\eg precision, recall, \etc) is often the endpoint of most contributions in the embryo assessment field. However, transforming a quantitatively performing model into a successful practical deployment is no easy task, esp. when considering the high-stakes involved in the context of IVF. The question on how to effectively and safely introduce this technology into a clinical setting where the different actors, including clinicians and patients, are involved, remains unsolved. Next we aim at advancing towards a solution, filling a gap in current literature.

\subsection{Opaque Models Create Responsibility Gaps}

The use of opaque AI models raises ethical and legal accountability concerns \cite{afnan_ethical_2021}, particularly when clinicians can not explain the decision-making process. This creates a ``responsibility gap", and without established accountability mechanisms it is challenging to determine who should be held responsible for any potential harm. For example, in cases of sub-optimal embryo selection or injury due to model recommendations, the decision-making process must be explainable to patients seeking to understand what happened or, in more extreme cases, seeking compensation for the damage caused. As of today, distrust in AI applications in medicine also comes from doctors’ fear of legal repercussions if something goes wrong due to unclear liability regimes \cite{panigutti2022understanding}. If clinicians base their decision on these opaque AI models, the evaluation of the decision-making process and, consequently, the determination of who is responsible will be greatly hampered.

It is then essential that the explanations provided by the model are as clear and understandable as possible by the clinician. For embryo image analysis this motivates the use of saliency maps obtained through feature attribution methods, which provide reliable (\ie true to the underlying model behavior) but also accessible evidence. To further ground the produced explanations in the context of clinical embryo selection, we suggest the mapping of this evidence-based support to the established annotation and evaluation guidelines and practices of the field, \eg through the association of feature attributions with the already established morphokinetic markers (such as a part of an image showing inappropriate cellular fragmentation), so that evidence provided is given in a language appropriate for the clinician and the problem. Still, due to the limitations of these XAI methods, it is recommended to adhere to ethical guidelines in AI, such as the one proposed by the EU \cite{lemonne_ethics_2018}. By following this framework, other key requirements such as data privacy and governance or human agency and oversight can be accomplished.

\subsection{Building Trust through Explanations}
The successful integration of AI-assisted embryo  selection systems in the medical context heavily depends on the acceptance by the clinicians and, consequently, their trust in them.

According to \cite{panigutti2022understanding}, there are several levels of trust which fall along a spectrum, ranging from complete distrust to over-reliance on AI systems.
Studies have shown that over-reliance on the suggestions can cause clinicians to take less initiative \cite{levy2021assessing} and also to be more likely to accept incorrect diagnosis \cite{harada2021effects}; this is known as \emph{automation bias} \cite{lee2004trust}. On the other side of the spectrum lie clinicians which do not trust an algorithm that they do not understand \cite{cai2019hello}, a phenomenon known as \emph{algorithmic aversion}.
As a result, not only training and education programs are necessary to ensure that clinicians understand the capabilities and limitations of AI-assisted embryo selection systems, but also making sure that the explanation or the model are not having negative impacts on the clinician's decision making process. 

It is recommended to avoid offering the discretized outcome of models (\eg this embryo is good) in favour of more detailed information, such as outcome probabilities (\eg this embryo is good with a probability of 55\%). For further bias avoidance, evidence may be presented only in the form of explanations, with the outcome hidden (\eg this area shows relevant visual features for determining if the embryo is good or not), so that the clinician interprets and integrates the visual features unbiasedly \wrt to the model's final prediction. Unfortunately, this approach is not without flaws either. It can induce \emph{confirmation bias} (\ie a clinician may only review the evidence that supports its own hypothesis), while preventing the methodology to exploit visual patterns not perceivable by humans, which may be a limiting factor in embryo analysis. The proposed solution is to provide all information available, outcome probabilities and saliency maps, and to extend that information to all relevant cases (\eg an area shows relevant visual features in favour of the embryo being good, an area shows features in favour of it being bad).



\subsection{Patient, Clinician, AI}

IVF is a highly personal procedure, with a large impact in the physical and mental health of individuals. Patients undergoing IVF have the right to know how decisions are being made about their own healthcare, and to maintain their decision-making power. To guarantee that, AI explanations should be also accessible to them; they cannot be left behind by the complexity of technology. Integrating explainability into AI-assisted embryo selection systems can not only target clinicians, it must also empower patients. 
By providing them with clear and easy to understand explanations of certain decisions, patients can be more involved and better equipped for the decision-making process regarding their own care. XAI results should be presented when relevant, and under the supervision of the clinician for its adequate interpretation and contextualization. Finally, it has to be mentioned that, since the use of these systems directly involves the processing of their data, the patient should be educated about how it will be processed and the use that will be made of it, in addition to asking for their informed consent.

\section{Conclusions}

We present an analysis of the presence of XAI for embryo selection on the literature. While there is a relative abundance of models for embryo selection and many have shown promising results when trained on retrospective data, the performance of these models in actual cases is yet unknown and their integration in the medical decision process remains unclear. Due to the opaque nature of these models, explainability is crucial to ensure their applicability. However, although some studies tackle the issue of explainability, the interpretations offered are often insufficient or do not rely on medical experts for evaluation. 
We enumerate a set of guidelines and suggestions that could help increase the interpretability and trustworthiness of these systems, with the goal of advancing towards the successful and safe integration of this technology in clinical practice.

\section*{Acknowledgements}
This work has received funding from the European Union’s Horizon 2020 research and innovation programme under grant agreement no.957331 - knowlEdge.


\printbibliography

\end{document}